1# Automated Gaming Pommerman: FFA

Members: Ms. Navya Singh, Mr. Anshul Dhull, Mr. Barath Mohan.S, Mr. Bhavish Pahwa, Ms. Komal Sharma

Bennett University, Greater Noida, India

Ns8558@bennett.edu.in, anshuldhull1997@gmail.com, barathmohansiva3@gmail.com, bhavishpahwa@gmail.com, komalsharma16dec@gmail.com## ABSTRACT-

Our game Pommerman is based on the console game Bommerman. The game starts on an 11 X 11 platform. Pommerman is a multi-agent environment and is made up of a set of different situations and contains four agents/players. Basically, they want other opponents to be an imperfect imitation of human-like behavior, and we will train our agent to outperform other players and optimize its score. We used Dijkstra's algorithm and tried implementing A* algorithm and the policies we implemented are DQN(Deep Q-Network), DQfD(Deep Q-learning from Demonstrations) and Imitation learning with Convolution 256. From these policies, the result of Imitation learning with Convolution 256 is observed to be the best.

**Keywords—** Reinforcement learning, Imitation learning, Deep Q learning from demonstrations, Pommerman## INTRODUCTION-

Reinforcement Learning is a sub-set of Machine Learning by which agents can become proficient in functioning by itself in the environment. Reinforcement Learning is one of the biggest trends in Data Science and is used in many areas like banking, healthcare, education and especially in gaming, robotics, and finance industry.[8] Reinforcement learning is better than ML algorithm because in ML we do not know what is the next step taken so it will not be effective, but in reinforced learning the Agent learns from the environment and thus this interaction with the environment helps it to achieve the next state using actions and for every action it achieves rewards which help it understand which action is useful and which is bad or unnecessary . As the rewards can be both positive and negative so the negative rewards help the Agent understand which actions are good or not. The player gets the initial state from the given environment and by the action performed by our Agent transitions to the first state for this, it gets the reward and evaluates the next action using this reward. The RL loop outputs a sequence of state, action, and reward.

**Multi-Agent Reinforcement Learning (MARL)**

Deep learning focuses on models that include multiple agents that learn by rapidly interacting with the environment. In many scenarios, agents in a MARL model can act cooperatively, competitively or exhibit neutral behaviors.[11] To tackle those complexities, MARL techniques borrow certain ideas from a game theory which can be very helpful when comes to model environments with multiple participants. Pommerman has four agents, power-ups, bombs galore in three modes which are: FFA, Team and Team Radio. In FFA four agents enter and one leaves. Planning, tactics with cunningness is tested in FFA and the board is fully observable.

**Challenges of MARL Agents:**

MARL models offer tangible benefits to deep learning tasks given that they are the closest representations of many cognitive activities in the real world.[6] However, there are plenty of challenges to consider when implementing this type of models. Without trying to provide an exhaustive list, there are three challenges that should be top of mind of any data scientists when considering implementing MARL models:

1. **The Curse of Dimensionality:** The challenge of deep learning systems is particularly relevant in MARL models. Many Multiagent Reinforcement Learning strategies work on certain game-environments which badly fail as the number of agents increase.
2. **Training:** Coordinating training across a large number of agents is another issue in MARL scenarios. These MARL models use training policy coordination mechanisms to minimize the impact of the training tasks.
3. **Ambiguity:** MARL models are very vulnerable to agent ambiguity scenarios. Imagine a multiplayer game in which two agents occupied the exact same position in the environment. To handle those challenges, the policy of each agent needs to consider the actions taken by enemies.

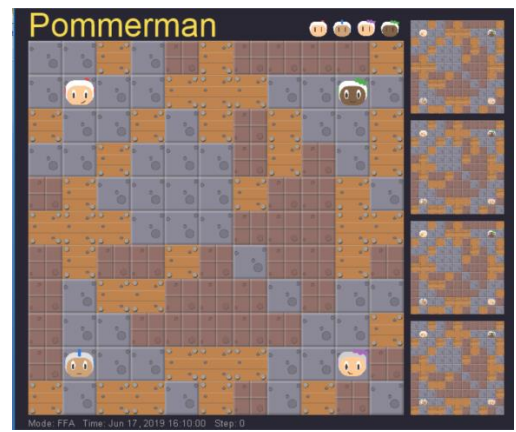

Figure 1: Pommerman environment at the starting of an episode.

The yellow ochre furnished wood and the brown ones are the rigid rough wall and the grey boxes are the open passages.

Rigid walls are perpetual and impenetrable. Wooden walls can be easily demolished by bombs. Unless they are blown up, they remain impenetrable. As they get destroyed they either turn into a passage or a power-up.[9]

On each and every step, the agent/player gets to choose one of the following actions:
1. Stop Action for a pass.
2. Up: Action for moving up on board.
3. Left: Action for moving left on board.
4. Down: Action for moving down on board.
5. Right: Action for moving right on board.
6. Bomb: This lays a bomb.

Overall observations received by the agent after each step are the following:
• Board (Size-121 Integer type): It is a flat board with a partially observable environment. The squares which are outside the 5x5 acumen around the agent's position are covered with fog's value (5).
• Position (Size-2 Integer type): It's the (x,y) position of the agent in the grid.
• Ammo (Size-1 Integer type): It's the present ammo of the agent.
• Blast Strength (Size-1 Integer type): It's the blast strength of the agent.
• Can Kick (Size-1 Integer type): Ability to kick the bomb by an agent.
• Teammate (Size-1 Integer type): It refers to agent's teammate.
• Enemies (Size-3 Integer type): It refers to the agent's enemy.
• Bomb Blast Strength (Integer type): The strength of bomb blast given for the bombs in the subview of the agent.
• Bomb Life (Integer type): The lifetime remaining for each bomb in the subview of the agent.
• Message (Size-2 Integer type): The message which is passed by the teammate.

Here the agent begins with one ammo (bomb) and it's ammo decreases by one whenever it plants a bomb. It's ammo increases by one when the bomb explodes. Blast strength of the agent at the start is 2. All the bombs it plants are imbued with the present strength of the blast. The life of the bomb is often timesteps. After termination of the bomb, it explodes and all the wooden walls, agents, powerups and other bombs within the range of its blast gets destroyed. Bombs which are destroyed in the way accumulate their explosion.

Power-ups: There are power-ups which are hidden in some of the wooden walls which are unraveled when these walls are destroyed. These are:
• Extra Bomb: This power-up increases the ammo of agent by one.
• Increase Range: This power-up increases the blast strength of the agent's bomb by one.
• Kick: This power-up helps the agent to kick the bomb. The kicked bomb travels in the direction that the agent was moving at unit timestep until they are encountered by the player, a bomb or a wall.

**POLICY**: It is a strategy which determines the way of agent's behavior at a particular time. It maps between the states taken into account and actions to be taken in those states.

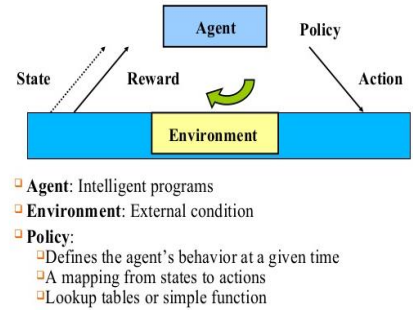

Figure 2: Elements of reinforcement learning

We suggest a given species of the policy gradient techniques for reinforcement learning whereas the given standard policy gradient techniques perform one gradient update at a time, we proposed an objective function that permits multiple epochs of minibatch updates. The new methods are known as **proximal policy optimization (PPO)** and have the perks of **TRPO**, but they are much easier to execute and are more general, have better sample complexity. The experiments we performed test PPO on a set of standard tasks, which includes the replication of robotic locomotion and Atari games, and we found that PPO surpasses all the other policy gradient techniques, and beats a distinct balance between given complexity, simplicity, and wall-time.

But we wanted to improve the efficiency and performance of our project so we tried implementing different policies such as DQfD(Deep Q-learning from Demonstrations)**,** A3C (Asynchronous Advantage Actor-Critic), DQN(Deep Q-Networks), DDPG(Deep Deterministic Policy Gradient). We implemented Dijkstra's algorithm for calculating the shortest path. We also tried to implement the A*(A star) algorithm as it is thought to be more efficient than Dijkstra's.

# RELATED WORK

Previously PPO policy has been used for training the agent.

**PPO:**
PPO stands for Proximal Policy Optimization it is quite well for environments which have a large number of state – action pairs as that of our Pommerman. We used tensor force to implement our PPO Agent and added a dense policy network.
So as to make our agent better we also chose Adam as the optimizer and learning rate at optimum. The Agent was trained by playing against three other Simple Agents and the results we obtained were in accordance with our expectations. We implemented DQfD policy instead of PPO as it was observed that DQfD is more efficient in training our agent.

**Learning from Demonstrations**: DQfD stands for Deep Q Learning from Demonstrations in this we try to pre train our Agent with Expert Demonstrations and then train the network on a mixture of new replay buffer and expert replay buffer. The denser the network the better our Agent gets. As we implemented DQfD from Tensor Force we had a small network and thus it's performance was limited but we still thought it could perform better than Simple Agent and would be able to demolish the Simple Agent.

But in DQfD our agent was playing defensive i.e. it was just trying to blow up the walls and gain power-ups rather than trying to attack the opponent players.

Thus we implemented imitation learning by collecting Reinforcement is used in many similar Atari games such as Mario, Frozen Lake, and Maze, etc.

A convolutional neural network is a branch of deep neural networks and most commonly used to analyze images. CNN's are regularized versions of multilayer perceptrons. Also cited to as Convolutional layer, it forms the groundwork of the CNN and functions the core operations of the training and therefore firing the neurons of the network. It performs the convolutional operation over the input volume as mentioned in the above section, and consists of a 3-D arrangement of the neurons. We used convolution 256.

**Mario:** The Machine learning program begins as a clean slate, not knowing anything of the task it is supposed to be doing. It then takes an input, which is the environment it has been placed in and starts trying to figure out which output gives it the highest rewards. This method is very much aligned with how humans learn to do new things through trial and error. We try things one way, then a different way, until we get the reward that we aim for. The main reason for this is that we include our past experiences in the game and can, therefore, better navigate the rules of the game to achieve a higher reward much faster. In machine learning, this would be considered training.

**Maze:** This has a Reinforcement learning methodology and the basic purpose is for the rat to get the maximum amount of cheese and at the same time try to save being eaten by the cat. The rat has to learn to explore the board effectively and at the same time maximize the reward which is the cheese.

## METHODOLOGY

We used the following methodologies in our project:

### ALGORITHMS:

1.) **Dijkstra's algorithm**- Dijkstra's Algorithm allows you to calculate the shortest path for the agent to move either to blow the wooden walls, escape from the flames or enemies, to gain powerups. It basically starts by considering path cost of those who are adjacent to the current node and the path value for all the other nodes as infinity and then keeps on updating the path value with the optimal value.

2.) **A\* algorithm**: A*(A-star) is analogous to Dijkstra algorithm but it is much more efficient than it. It uses a heuristic value for determining the cost of the path, so it only considers the most optimal path out of every path. Its function uses g(n) and h(n) values for calculating f(n).

### POLICIES:

**DQN**- In Deep Q-Learning or DQN for each given input state, the network outputs predicted Q-values for each action that can be taken from that particular state. The objective of this network is to find the optimal Q-function, and the optimal Q-function will satisfy the Bellman equation.

Then, the loss in the network is determined by correlating the outputted Q-values to the target Q-values from the right-hand side of the Bellman equation, and like with any other neural networks, the objective here is to minimize this loss.

After the loss is calculated, the weights within the network are updated via backpropagation and SGD and just like with any other typical neural network. This process is done over and over again for each and every state in the environment until we adequately curtail the loss and get an approximate optimal Q-function.

In a complex environment like video games, we input images(still frames) that capture states from the environment as the input to the network.

**DQfD**- Deep reinforcement learning algorithms have achieved remarkable results on a number of problems once thought to be unsolvable without the aid of human Integer intuition and creativity. Reinforcement Learning agents can learn to ace tasks like chess and retro video games without any prior instruction—often overpowering the performance of even the greatest human experts. These methods are sample inefficient and rely on learning from hundreds or even thousands of complete failures before any progress is made. That's a luxury we can afford when the task is simple or can be simulated, like an Atari screen or chess board, but is at least partially responsible for RL's relatively short list of real-world applications.

**Imitation(Deep Reinforcement Learning)**- We used a python script to render our Pommerman environment and run 600 episodes to gather around 600K observations of Simple Agent and collected a Dataset of the same. Then we made a model and tried to train our model using the Dataset and implemented our model on an Agent and called it the Imitation Agent and then to evaluate our Agent rendered a battle with three Simple Agents.

## EXPERIMENTAL RESULTS-

**Platforms used:**

- Windows 10
- Pycharm 2019.1.3
- Jupyter Notebook
- Google Colab
- Kaggle

**DQN**-

We implemented the DQN policy on our agent using the tensor force library to import the DQN Agent. Using the Q learning model we imported and implemented the DQN agent as our Pommerman agent and observed the effect by running around 5 episodes with 3 other Simple Agents:-

> Stats: Episode Rewards: [-1, -1, -1, -1, -1]
> Episode Time:
> [6.79481,1.57478,2.07544,0.53315,0.85172]

Figure 3: DQN output

The Agent learns that it has to prevent itself from bombs and that the bombs are dangerous for itself and finally it adopts a defensive technique and instead of placing bombs it tries to find a corner between any two walls to protect itself, waiting for the other agents to eliminate themselves. It also tries to prevent from other incoming agents.

**DQfD**-

After implementing DQN Agent we also tried to implement DQFD Agent because it is thought to be more efficient than DQN because it uses demonstrations or so-called observations on DQN. We implemented the DQFD agent in for our Pommerman environment and observed by running around 5 episodes with 3 other Simple Agents:-

> Stats: Episode Rewards: [-1,-1,-1,0,-1]
> Episode Time:
> [9.08531,5.71206,11.1065,14.07491,7.284523]

Figure 4: DQfD output

The DQFD Agent tries to explore or move through the board and also learns to effectively place bombs and tries to evade enemies but still cannot manage to stand till the end however is always the second last left in battle and on an average manages to eliminate one Simple Agent.

Imitation learning with Convolution 256-

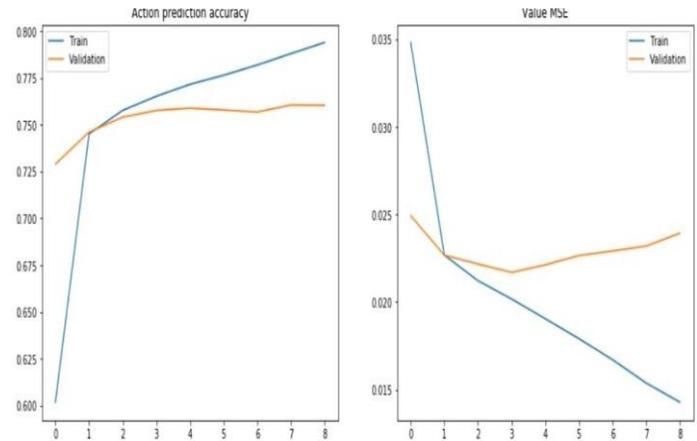

Figure 5: Imitation learning output

## CONCLUSION-

In our Pommerman emulation, we try to implement as many policies as we can to make our Agent compete with the Simple Agent which is designed by the creators of the Pommerman Competition. As the creators ensure us that we can use any method to train our Agent and they would host it without any limitation at all we tried various different policies and methodologies. First, we try Proximal Policy Optimization (PPO) as it is given as a sample experiment by the creators to learn about various other policies through its implementation. Then we move to DQN because of the successful results we found it had performed in the past. Implementation of DQN leads us to the result that in DQN the Agent becomes defensive and tries to hide in a corner or in between walls and does not explore the play area and place bombs. The results of DQN indicated us to move to DQfD as it is better suited for Pommerman experiment as in DQfD we have a set of Human Demonstrations and it can help our Agent train for exploration. The results of DQfD were up to our expectations and the Agent learned to explore and moved through the board and placed bombssuccessfully. Then we tried imitation learning using Simple Agent as our Model Agent and had a breakthrough. The imitation learning Agent matches the performance level of Simple Agents and eliminates Simple Agents and is the last one left standing around 25% of all episodes. Thus we achieved to make an Agent powerful enough to eliminate Simple Agents and also save itself from them.